%% file: main.tex
\documentclass[acmsmall]{acmart}

\AtBeginDocument{%
  }

\setcopyright{acmcopyright}
\copyrightyear{2024}
\acmYear{2024}
\acmDOI{XXXXXXX.XXXXXXX}

\acmJournal{JACM}
\acmVolume{37}
\acmNumber{4}
\acmArticle{111}
\acmMonth{8}

\usepackage{enumitem}
\usepackage{bm}
\usepackage{amsmath}

\usepackage{amsthm}
\usepackage{amssymb}
\usepackage{graphicx}
\usepackage[linesnumbered,ruled,vlined]{algorithm2e}
\usepackage{float}
\usepackage{booktabs}
\usepackage{subfigure}
\usepackage{makecell}
\usepackage{multirow}
\usepackage{lineno}
\usepackage{siunitx}
\usepackage{url}
\usepackage{dsfont}
\usepackage{xcolor} 

\usepackage{newfloat}
\usepackage{listings}
\usepackage{pifont}
\newcommand{\figref}[1]{Fig. \ref{#1}}
\newcommand{\tabref}[1]{Table \ref{#1}}
\newcommand{\secref}[1]{Section \ref{#1}}
\newcommand{\equref}[1]{Equation (\ref{#1})}

\newcommand{\myref}[1]{(\ref{#1})}

\begin{document}

\title{Adaptive Taxonomy Learning and Historical Patterns Modelling
for Patent Classification}

\author{Tao Zou}
\email{zoutao@buaa.edu.cn}

\author{Le Yu}
\email{yule@buaa.edu.cn}

\affiliation{%
  \institution{State Key Laboratory of Complex \& Critical Software Environment, Beihang University}
  \city{Beijing}
  \country{China}
  \postcode{100191}
}

\author{Junchen Ye}
\email{junchenye@buaa.edu.cn}

\affiliation{%
  \institution{School of Transportation Science and Engineering, Beihang University}
  \city{Beijing}
  \country{China}
  \postcode{100191}
}

\author{Leilei Sun}
\authornote{Corresponding Author}
\email{leileisun@buaa.edu.cn}

\affiliation{%
  \institution{State Key Laboratory of Complex \& Critical Software Environment, Beihang University}
  \city{Beijing}
  \country{China}
  \postcode{100191}
}

\author{Bowen Du}
\email{dubowen@buaa.edu.cn}

\affiliation{
  \institution{School of Transportation Science and Engineering, Beihang University}
  \city{Beijing}
  \country{China}
}

\affiliation{%
  \institution{State Key Laboratory of Complex \& Critical Software Environment, Beihang University}
  \city{Beijing}
  \country{China}
  \postcode{100191}
}

\affiliation{
  \institution{Zhongguancun Laboratory}
  \city{Beijing}
  \country{China}
}

\author{Deqing Wang}
\email{dqwang@buaa.edu.cn}

\affiliation{%
  \institution{State Key Laboratory of Complex \& Critical Software Environment, Beihang University}
  \city{Beijing}
  \country{China}
  \postcode{100191}
}


\renewcommand{\shortauthors}{Zou et al.}

\begin{abstract}
Patent classification aims to assign multiple International Patent Classification (IPC) codes to a given patent. 
Existing methods for automated patent classification primarily focus on analyzing the text descriptions of patents. However, apart from the textual information, each patent is also associated with some assignees, and the knowledge of their previously applied patents can often be valuable for accurate classification. Furthermore, the hierarchical taxonomy defined by the IPC system provides crucial contextual information and enables models to leverage the correlations between IPC codes for improved classification accuracy. However, existing methods fail to incorporate the above aspects and lead to reduced performance.
To address these limitations, we propose an integrated framework that comprehensively considers patent-related information for patent classification. To be specific, we first present an IPC codes correlations learning module to capture both horizontal and vertical information within the IPC codes. This module effectively captures the correlations by adaptively exchanging and aggregating messages among IPC codes at the same level (horizontal information) and from both parent and children codes (vertical information), which allows for a comprehensive integration of knowledge and relationships within the IPC hierarchical taxonomy. Additionally, we design a historical application patterns learning component to incorporate previous patents of the corresponding assignee by aggregating high-order temporal information via a dual-channel graph neural network.
Finally, our approach combines the contextual information from patent texts, which encompasses the semantics of IPC codes, with assignees' sequential preferences to make predictions. Experimental evaluations on real-world datasets demonstrate the superiority of our proposed approach over existing methods. Moreover, we present the model's ability to capture the temporal patterns of assignees and the semantic dependencies among IPC codes.
\end{abstract}




\maketitle
\input{section-1-introduction}

\input{section-3-related-work}

\input{section-2-preliminaries}

\input{section-4-methodology}

\input{section-5-experiments}

\input{section-6-conclusion}

\begin{acks}
This work was supported by the National Natural Science Foundation of China (62272023, 62276015, 51991391, 51991395) and the S\&T Program of Hebei (225A0802D).
\end{acks}

\bibliographystyle{ACM-Reference-Format}
\bibliography{reference}

\end{document}

%% file: section-1-introduction.tex
\section{Introduction}
\label{section-1}
Patent classification is essential for the management of patents, which aims to assign multiple International Patent Classification (IPC) codes to a given patent based on the IPC taxonomy system\footnote{\url{https://ipcpub.wipo.int}}. Traditional classifications of patents heavily rely on manual efforts and domain knowledge, which are expensive and become infeasible due to the recent surge in the number of patents (e.g., 3.40 million in 2021\footnote{\url{https://www3.wipo.int/ipstats/keysearch.htm?keyId=201}}). Therefore, automatically making accurate classifications of patents is urgently needed.

\begin{figure}[!htbp]
    \centering
    \includegraphics[width=1.0\columnwidth]{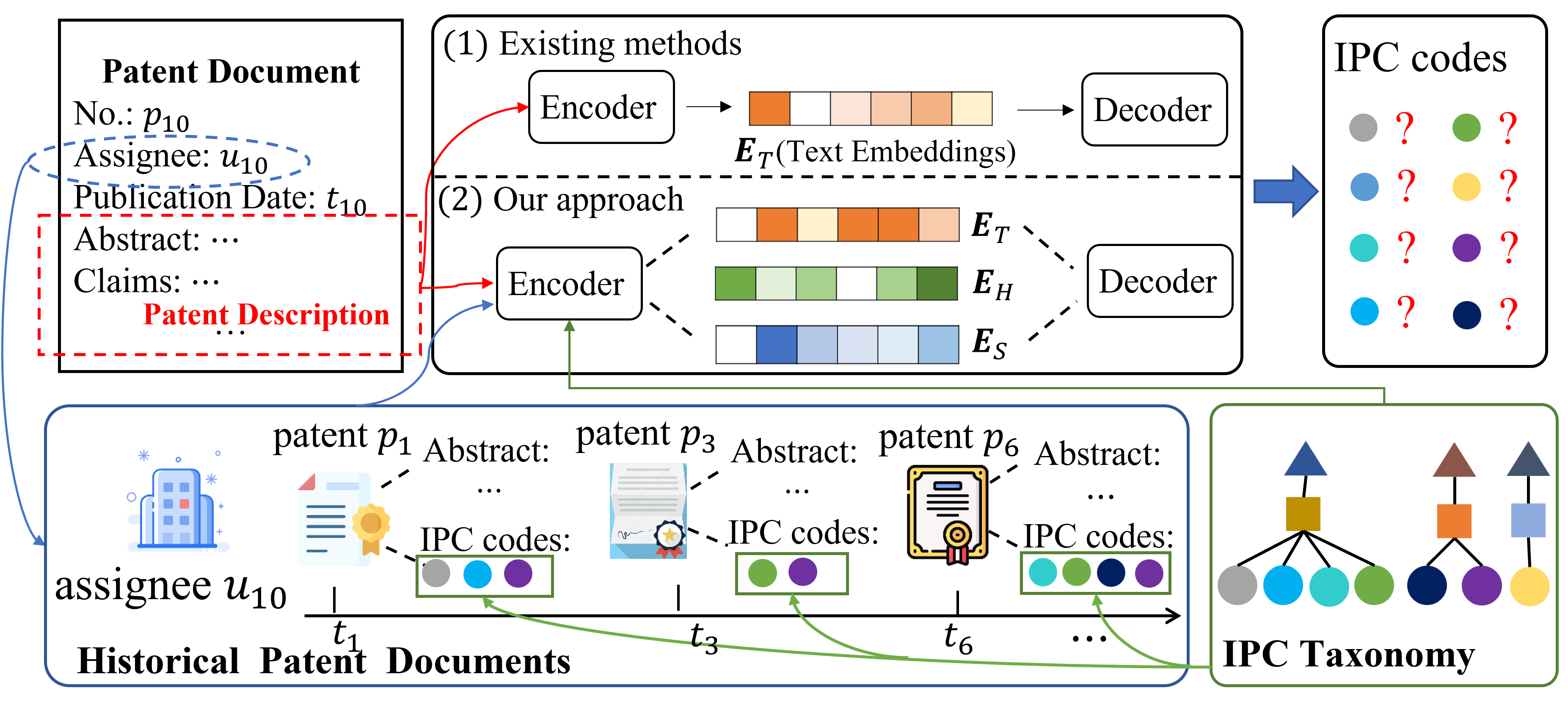}
    \caption{The task is to predict IPC codes of patent documents. We show the difference between existing works and our framework. Specifically, $\bm{E}_T$ represents the textual embeddings of patent descriptions, $\bm{E}_H$ denotes the semantic embeddings of classification codes in the hierarchy taxonomy and $\bm{E}_S$ is the temporal representation of assignee $u_{10}$ learned from historical patents.}
    \label{fig:motivation}
\end{figure}

In recent years, several methods have been proposed to solve the patent classification problem. One part of the researchers attempted to analyze the patent documents with feature engineering. For example, \citet{DBLP:journals/entropy/HuLYYYH18} utilized patent text information to exact representative keywords for classification while \citet{DBLP:conf/jcdl/LiCZL07} analyzed patent citation network for classification. Another part of existing works employed machine learning methods or deep learning algorithms to capture hidden relationships in patents. \citet{DBLP:conf/bigcomp/RoudsariALL20} utilized BERT pre-trained model to capture informative textual information for classification and \citet{DBLP:journals/scientometrics/LiHCH18} applied CNN algorithm to capture the hidden relationships in text sequence. \citet{DBLP:conf/aaai/Tang0XPWC20} learned the correlations between words in text descriptions via the attention mechanism and GCN framework for patent classification. However, most existing methods primarily focus on analyzing the textual information within patent documents for classification purposes.

\begin{figure}[!htbp]
    \centering
    \includegraphics[width=1.0\columnwidth]{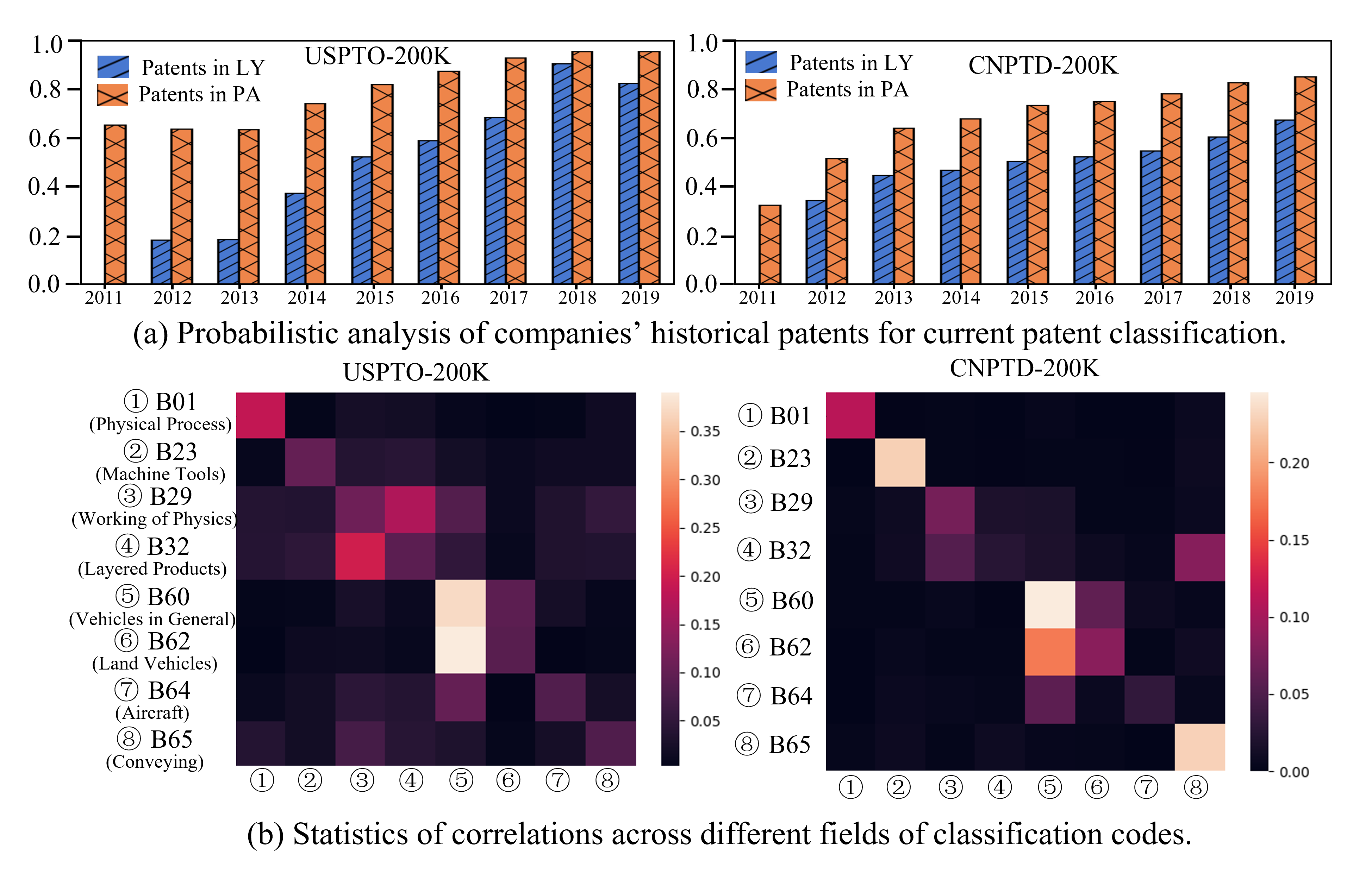}
    \caption{In scenario (a), we analyze assignees' repeated application behaviors by calculating the recurrence rates of IPC codes in current patents compared to historical patents applied in the last year or up to the present time. In (b), we depict the average co-occurrence ratios of different IPC codes assigned in patents in the ``Transportation" field in two datasets and normalize the performance in each row.}
    \label{fig:introduction}
\end{figure}

In practical scenarios, it is essential to consider not only the textual information but also the recurring application behaviors demonstrated by assignees over time. It is worth noting that assignees often show a tendency to apply for specific patents within their respective fields. To verify this point, we analyze by calculating the co-occurrence ratios of identical IPC codes for each assignee in both current and historical patents within the last year (referred to as LY) or in the past (i.e., encompassing data up to the current patent, referred to as PA). The averaged results for all assignees are presented in \figref{fig:introduction} (a) \footnote{Please refer to \secref{section-4} for detailed dataset information.}. Notably, our findings reveal significant correlations between historical patent applications and the classification of IPC codes in current patents, indicating the possible existence of sequential application behaviors among assignees. Consequently, the observations suggest that incorporating assignees' sequential behavior patterns has the potential to enhance the accuracy of patent classification.

Moreover, patent classification can benefit from incorporating the semantic dependencies among IPC codes, which are organized in a hierarchical taxonomy. To explore this aspect, we compute the number of patents assigned to specific subdomains within the ``Transportation" field and normalize the results. This analysis is depicted in \figref{fig:introduction} (b). Specifically, the diagonal line denotes the proportion of patents in each field, while each row signifies the ratio of patents that are simultaneously assigned with IPC codes from other fields. We discover distinct semantic dependencies among IPC codes in certain fields, such as ``B60" and ``B62" in the USPTO-200K and CNPTD-200K datasets. These dependencies indicate inherent relationships between technological areas. Furthermore, we observe that these semantic relationships vary across different fields within the two datasets, highlighting the importance of capturing adaptive semantic correlations among IPC codes within the taxonomy. By considering and incorporating these adaptive semantic correlations among IPC codes, we can overcome the limitations of conventional methods that solely focus on learning the fixed structure of the taxonomy \cite{DBLP:conf/ijcai/LuRCKN21,DBLP:conf/ijcai/ShangMXS19} and provide a more robust and efficient solution for patent classification.

Building upon these findings, we propose an integrated framework for patent classification, which leverages the semantic information. Our approach combines adaptively taxonomy learning and historical application modeling to achieve enhanced classification accuracy. To provide a clear conceptual comparison with existing works, we present a visual representation in \figref{fig:motivation}.
In particular, our model is designed to incorporate various information on patents, including the text description of patents, the IPC taxonomy system, and the assignees' historical patent records. Firstly, we adopt a text embedding encoder to represent the textual information of patents effectively. Secondly, considering the semantic dependencies of IPC codes within the taxonomy, we adaptively learn the dependencies across IPC codes at both vertical and horizontal levels using an attention mechanism on the taxonomy system, which allows us to capture the semantic belongingness of IPC codes accurately. Thirdly, we construct a weighted directed graph for each assignee based on their historical patent sequence. By aggregating messages from historical patents within the graph, we learn the high-order complex application behaviors of assignees. Additionally, in order to capture the temporal behavior patterns explicitly, we utilize positional embeddings for each patent in the graph. At last, we integrate the semantic correlations among IPC codes into the context embeddings of patents and combine the sequential behaviors information of assignees for patent classification. Extensive experiments on two real-world datasets show that our model outperforms existing methods. In addition, we further analyze the importance of learning semantics among IPC codes and sequential information of assignees on patent classification. We summarize the main contributions of this paper as follows.
\begin{itemize}

    \item 
    A historical application patterns learning component is proposed. Unlike the existing works that utilized sequential networks to learn assignees’ temporal applying preferences, we build a patent interaction network among their historical patents with a sliding window strategy and model the high-order applying behaviors through dual message aggregation in the graph neural network.

    \item 
    An IPC codes correlations learning module is designed. We not only capture the bi-directional relationships for classification codes along with their parent and children in the hierarchy but also learn the adaptive semantic correlations between classification codes at the same level simultaneously while adhering to hierarchy constraints.

    \item 
    We have gathered two comprehensive datasets for patent classification in multiple research scenarios. These datasets contain abundant metadata information that has been previously overlooked, including applicants, assignees, and publication time. All the codes and datasets are publicly available online\footnote{https://github.com/Hope-Rita/PatCLS}, and we hope that it could facilitate further research and advancements in this field.
\end{itemize}

%% file: section-3-related-work.tex
\section{Related work}
\label{section-5}
\subsection{Patent Classification}
Patent classification aims to classify patents into different categories based on their technological information, which helps researchers and companies to search for and retrieve patents efficiently. Existing methods can be divided into three categories. The first category classified the patents with feature engineering. For instance, \citet{DBLP:journals/entropy/HuLYYYH18} distinguished the representative keywords as the categories of patent documents, and \citet{DBLP:conf/jcdl/LiCZL07} aggregated the labels of each patent's references in the patent citation network for classification. However, these methods may not capture the complex and informative semantics present in patents as the dataset size increases. The second category used traditional machine learning such as Support Vector Machine (SVM) \cite{DBLP:journals/asc/WuKH10}, k-Nearest Neighbors (k-NN) \cite{DBLP:journals/jasis/LiuS11}, and Artificial Neural Networks (ANNs) \cite{DBLP:journals/candie/LeeLY12}. For example, \citet{DBLP:journals/asc/WuKH10} proposed a hybrid genetic Support Vector Machine (HGA-SVM) to automatically classify patent texts by learning the experts' knowledge, while \citet{DBLP:journals/candie/LeeLY12} designed a three-phase framework, including SVM, K-NN, and K-means for patent classification. However, as the scale of the datasets increased, traditional machine learning methods were unable to capture the informative and complex semantics present in patents.
The third category employed deep learning methods to capture the textual information from patents, such as LSTM (Long Short-Term Memory) \cite{DBLP:journals/scientometrics/LiHCH18, DBLP:conf/sdm/ShalabySSG18}, BERT \cite{DBLP:conf/bigcomp/RoudsariALL20, DBLP:journals/scientometrics/RoudsariALL22}, CNN (Convolutional Neural Network) \cite{hu2018hierarchical, DBLP:conf/pkdd/AbdelgawadKGFH19} and GNN (Graph Neural Network) \cite{DBLP:conf/aaai/Tang0XPWC20, DBLP:journals/www/FangZWXZC21}. For instance, \citet{DBLP:journals/scientometrics/LiHCH18} combined word embedding and CNN algorithm for patent classification while \citet{DBLP:conf/bigcomp/RoudsariALL20} employed Bert \cite{DBLP:conf/naacl/DevlinCLT19} model and used fine-tuning approach in patent classification. \citet{DBLP:conf/aaai/Tang0XPWC20} utilized attention mechanism and GCN framework for multi-label classification. 

Our work goes beyond simply capturing textual information from patent documents. Instead, it proposes a method to learn the representations of patents by integrating the hierarchical semantic structural dependencies across IPC codes in the taxonomy and the assignees' temporal preferences for applying to patent documents for patent classification. This approach allows for a more comprehensive understanding of the complex relationships inherent in patent documents, resulting in more accurate and effective classification.

\subsection{Multi-label Textual Classification} 
The purpose of multi-label textual classification is to predict the set of labels for a given text document. In recent years, lots of studies have been proposed for learning textual information. Some methods utilized traditional machine learning algorithms for prediction. For example, one of the popular approaches treated the problem as a binary classification for each label, which trained separate classifiers for labels and predicted the final results with some aggregation methods \cite{DBLP:conf/icml/YenHRZD16, DBLP:conf/wsdm/BabbarS17}, while another branch utilized a decision tree algorithm and built a decision tree for labels based on text data \cite{DBLP:conf/kdd/PrabhuV14, DBLP:conf/www/PrabhuKHAV18}. Recently, some models applied deep learning methods to solve MTC tasks. \citet{DBLP:conf/aaai/YehWKW17} combined canonical correlation analysis and an auto-encoder model to capture label dependencies, while TextGCN \cite{DBLP:conf/aaai/YaoM019} built a text graph to connect words and documents then embedded words and documents together based on GCN framework. Besides, some studies \cite{DBLP:conf/cikm/HuangCLCHLZZW19, DBLP:conf/ijcai/ShangMXS19, DBLP:conf/acl/WangZLCZW20, DBLP:conf/emnlp/XuTZGXJR21, DBLP:conf/naacl/ShenQMSRH21, DBLP:conf/ijcai/LuRCKN21} figured out that labels are organized as a hierarchical tree in some scenarios and utilized the hierarchical structure within the label taxonomy for text classification at each level. For example, 
\citet{DBLP:conf/emnlp/XuTZGXJR21} pre-defined a fixed correlation matrix between classification codes and aggregated semantic information along the hierarchical taxonomy and \citet{DBLP:conf/naacl/ShenQMSRH21} leveraged only class names as supervision signals to predict the core classes and their ancestors via a top-down selection strategy. \citet{huang2022hmcnet} learned the implicit dependencies across classification codes at adjacent levels with a recurrent-based network and generated incoherent predictions via a prune-based path selection strategy in the hierarchy. Besides, \citet{amigo2022evaluating} defined an information-theoretic-based metric for hierarchical multi-label classification based on the unbalanced distribution between classes.

Although existing works have attempted to capture hierarchical dependencies among labels, most of them just learned the fixed relationships along classification taxonomies, which prevents the model from learning complex structural dependencies across different labels. In this work, we model the adaptive inherent semantics across labels at the same level or across different levels while incorporating the constraint of hierarchical taxonomy to learn abundant structural information. Furthermore, in contrast to previous methods, this approach integrates bi-directional vertical information along the taxonomy, helping to capture the consistent hierarchical semantics.

\subsection{Temporal Behavior Modeling}
The goal of temporal behavior modeling is to capture the dynamics and patterns of temporal data for predicting the future behavior of targets \cite{DBLP:journals/tois/FangZSG20}. Recently, many methods have been proposed in various domains. For instance, \citet{DBLP:journals/corr/abs-2205-14837} and \citet{DBLP:conf/aaai/MaMZSLC20} applied GNN based on temporal interactions to embed global context information and capture the long-term and short-term behaviors of items respectively in recommendation system. \citet{DBLP:conf/sigir/0001DWLZ020} effectively propagated message information between users and items in the interaction graph without using feature transformation and nonlinear activation. 
\citet{DBLP:conf/kdd/YuSDL0L20} and \citet{DBLP:conf/www/YuWS0L22} leveraged dynamic GNNs to solve the temporal sets prediction problem, which is a general formalization of the next-basket recommendation task \cite{DBLP:conf/www/RendleFS10}.
To capture the sequential behaviors of patients in intelligent treatment, 
\citet{DBLP:conf/kdd/WangZHZ18} and \citet{DBLP:conf/kdd/0001YSLQT18} employed the recurrent neural network with supervised reinforcement learning, and a heterogeneous long short-term network framework respectively. In social networks, \citet{DBLP:journals/tnn/JiaoGJHWPGW22} combined both network topology and temporal information to predict link relations based on deep autoencoders, while \citet{DBLP:journals/tkde/QiuWHDYY23} represented motifs in social networks as temporal distributions and made predictions based on a combination of multiple motifs. In patent application trend prediction, \citet{DBLP:journals/tkde/ZouYSDWZ24} designed memorable representations for companies and IPC codes for learning their evolving semantic information based on historical patent sequences.

The existing works are proposed for learning specific dynamics in several scenarios, which are not specifically designed for patent classification problems. In this paper, we build a patent interaction graph from the historical patent sequences and design a historical application patterns learning module to model the high-order temporal application behaviors of assignees based on their previous patent records via graph neural networks with positional embedding.

%% file: section-2-preliminaries.tex
\section{Preliminaries}
\label{sec:section-2}
This section first presents the related concepts about patent classification and then formalizes the studied problem.

\subsection{Definitions}

\textbf{Hierarchical Classification Taxonomy.}
It consists of all the available IPC codes organized in a hierarchical structure. We use $\mathcal{T}=\{\mathcal{C},\mathcal{R}\}$ to represent the IPC taxonomy, where $C^l$ represents the IPC codes in the $l$-th level of IPC taxonomy, and $\mathcal{R}=\{\bm{R}^{1,2},\cdots,\bm{R}^{L-1, L}\}$ denotes hierarchical relationships of the codes. $\bm{R}^{l,l+1}\in\mathbb{R}^{|C^{l}|\times|C^{l+1}|}$ describes the belongingness of $(l+1)$-th level codes to $l$-th level codes. $\bm{R}^{l,l+1}_{i,j}=1$, if $c^l_i$ is the parent code of $c^{l+1}_j$; $\bm{R}^{l,l+1}_{i,j}=0$, otherwise. To further explore the hierarchical structure, we denote that $S^{H}_{c^{q}_{i}}$ as the set of IPC codes at the $q$-th level that shares the same parent of $c^q_i$, which helps us gain insights into related fields in the taxonomy. On the other hand, we define $S^{V}_{c^{q}_{i}}$ as the set comprising both the parent of $c^q_i$ at the $q-1$-th level and the children of $c^q_i$ at the $q+1$-th level, which allows us to learn the immediate relationships surrounding a specific code.

\textbf{Patent Document.} A patent document is associated with a unique patent number, which also includes patent assignee, title, publication time, IPC codes, and other information. In this paper, a patent with ID $k$ is represented as $p_k = (W_k, u_k, t_k, \mathcal{Y}_{k})$, where $W_k$ is the text description of patent document $p_k$, $u_k$ is the assignee, $t_k$ is the publication time, and $\mathcal{Y}_{k} = \{{\mathcal{Y}}_k^1,\cdots,{\mathcal{Y}}_k^L\}$ is the set of assigned IPC codes at different levels in the classification taxonomy $\mathcal{T}$, where ${\mathcal{Y}_k^l} \subset C^l$ are at the $l$-th level.

\subsection{Problem Formalization}
Following \cite{DBLP:conf/aaai/Tang0XPWC20, DBLP:conf/cikm/HuangCLCHLZZW19}, the target of this paper is to develop an automatic procedure that can assign a candidate patent $p_k$ with a few IPC codes at a specific level according to all the available information, which enables classification at the desired level of granularity. This task could be formalized as, $$\hat{\bm{Y}}_{k} = f\left(W_k, \mathcal{T}, P_{u_k}^{t_k}\right),$$
where $W_k$ is the description text of this patent $p_k$, $\mathcal{T}$ is the IPC taxonomy system, and $P_{u_k}^{t_k}=\{p_i|u_i=u_k, 0 \leq t_i\leq t_k\}$ represents the historical patent applying records of assignee $u_k$ prior to time $t_k$. $\hat{\bm{Y}}_{k}$ represents the predicted probability of all IPC code at the $q$-th level. Here, $q\in [1, \cdots, L]$ represents a specific level within the IPC taxonomy hierarchy. 

The assignment of IPC codes not only depends on the patent description but also requires modeling assignees' preferences based on their historical application records. Additionally, capturing the underlying semantics of the taxonomy holds significant importance. To address these challenges, this paper presents a comprehensive framework for tackling the patent classification problem. The subsequent sections elaborate on the various modules designed to leverage the aforementioned information and enhance the accuracy of the classification.

%% file: section-4-methodology.tex
\section{Methodology}
\label{section-3}

\begin{figure*}[!htbp]
    \centering
    \includegraphics[width=1\columnwidth]{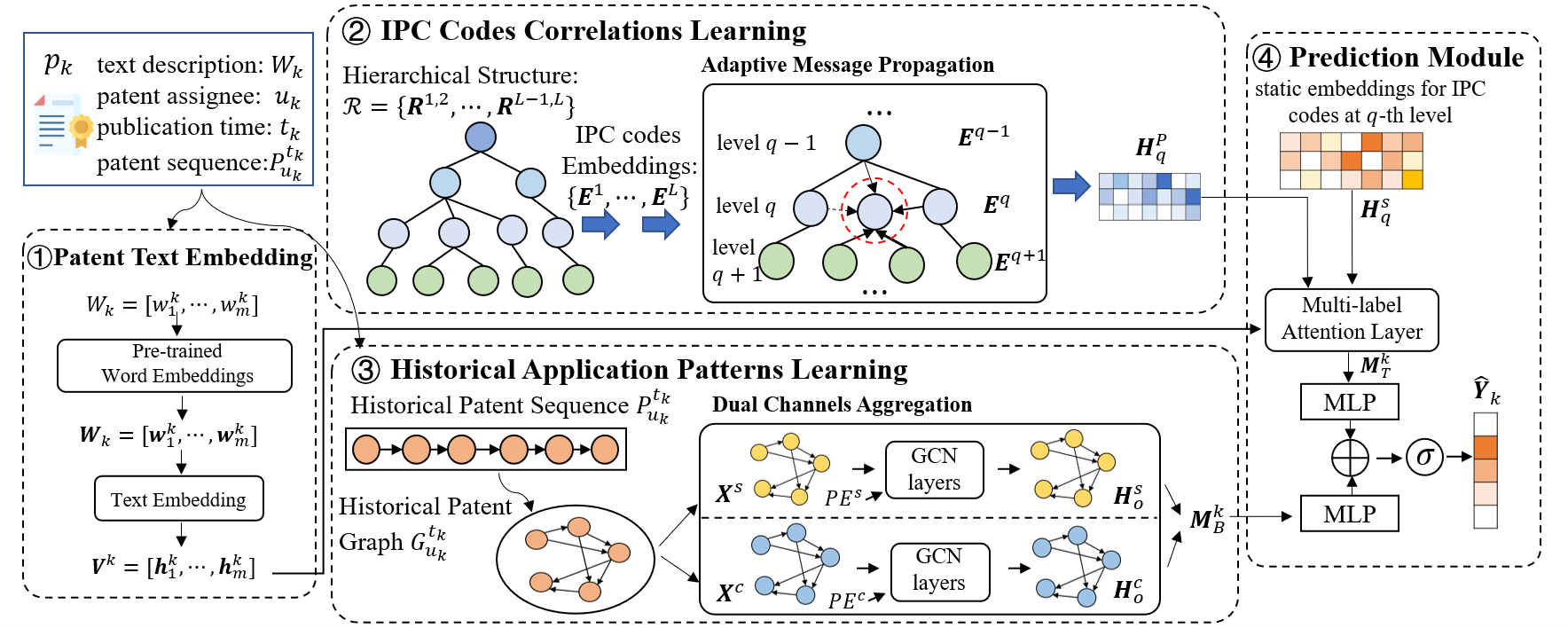}
    \caption{Framework of the proposed model.}
    \label{fig:framework}
\end{figure*}

This section first presents the framework of the proposed model and then provides details of each module. Illustrated in \figref{fig:framework}, our model consists of four key components: \ding{172} Patent text embedding module, which converts the textual description of a patent into a semantic vector, enabling meaningful contextual representations. \ding{173} IPC codes correlations learning module, which establishes the semantic relationships among IPC codes via adaptively propagating messages along the IPC taxonomy at the vertical and horizontal levels. \ding{174} Historical application patterns learning module, that learns the high-order temporal preferences of assignees by aggregating contextual and label information simultaneously from their historical patents. \ding{175} Prediction module aims to predict the classification probabilities of patent documents by decoding the information from contextual embeddings, which incorporate the semantics of IPC codes, as well as the application preferences of assignees.

\subsection{Patent Text Embedding}
Text representation learning has been widely studied in the field of Natural Language Processing (NLP). As a result, a number of methods have been proposed, such as Bi-LSTM \cite{DBLP:journals/corr/Graves13, DBLP:journals/neco/HochreiterS97}, CNN \cite{DBLP:journals/corr/OSheaN15}, Transformer \cite{DBLP:conf/nips/VaswaniSPUJGKP17}, BERT \cite{DBLP:conf/naacl/DevlinCLT19}, GPT-3 \cite{DBLP:conf/nips/BrownMRSKDNSSAA20}, etc. Moreover, most of these methods have been employed in Extreme Multi-label classification (XML) or patent classification problems. In this paper, our focus lies on learning IPC code semantics from the IPC taxonomy and capturing the assignee preferences from historical patent records. Consequently, we employ a simple text embedding approach to encode the patent description text. 
For a patent $p_k$, we first select $N$ key words from its description text, denoted as $W_k=\left[w^k_1, w^k_2, \cdots, w^k_N\right]$. Then, each word is represented by a $T$-dimensional vector, obtained from a pre-trained model. Hence, the description text could be represented as $\bm{W}_k=\left[\bm{w}^k_1, \bm{w}^k_2, \cdots, \bm{w}^k_N\right]$. To incorporate the contextual information for each word, we utilize a Bi-LSTM mechanism \cite{DBLP:journals/corr/Graves13} to generate the final patent text embedding $\bm{V}^k \in \mathbb{R}^{N\times 2F}$.

\subsection{IPC Codes Correlations Learning}
As stated in the Preliminaries Section, IPC codes are organized hierarchically within a taxonomy system\footnote{\url{https://ipcpub.wipo.int}} that categorizes them according to the semantic relationships, which allows us to explore and learn the inherent semantic correlations across IPC codes. In this work, we design an IPC codes correlations learning module to leverage this hierarchical organization. This module aims to capture the hidden dependencies between IPC codes by aggregating their embeddings horizontally and vertically along the taxonomy. Specifically, we define the embeddings of IPC codes within the hierarchical taxonomy as $\{\bm{E}^1,\cdots,\bm{E}^L\}$, where $\bm{E}^l\in \mathbb{R}^{|C^l|\times2F}$ represents the embeddings for IPC codes at the $l$-th level. Firstly, we adaptively propagate the semantic information of IPC codes at each level, considering both vertical and horizontal relationships. Subsequently, we aggregate the embeddings of IPC codes hierarchically throughout the taxonomy to generate comprehensive representations that incorporate global and local information for IPC codes.

\textbf{Adaptive Message Propagation.} The taxonomy system provides information about the hierarchical relationships between IPC codes at adjacent levels. Previous studies \cite{DBLP:conf/ijcai/LuRCKN21,DBLP:conf/ijcai/ShangMXS19} have focused on capturing fixed semantic dependencies along the taxonomy by aggregating embeddings from IPC codes at higher levels. However, in addition to the semantics at higher levels, valuable information can be learned from IPC codes at the same or lower levels. Moreover, the fixed structure of aggregation may limit the ability of models to learn complex knowledge across IPC codes. To address these limitations, we propose adaptive message propagation to capture detailed relationships between IPC codes at adjacent levels. 

In particular, we aggregate information for each IPC code $c^q_i$ from two perspectives. Firstly, at the horizontal level, we calculate the similarity between IPC codes at the same level and then adaptively propagate the information among IPC codes that share the same parents via an attention mechanism. This allows us to capture correlations in the related fields, providing a more comprehensive understanding of the interconnections between IPC codes. Secondly, we extend this message propagation to the vertical level. We aggregate information from adjacent levels, including the parent codes at the higher level $q-1$ and children at the lower level $q+1$ along the taxonomy. Hence, we can learn the adjacent relationships surrounding IPC codes. The process can be described as follows,
\begin{equation}
\label{equ:new hierarchy_horizon}
\begin{split}    
    \beta_{i,j}^{q,p} &= \bm{E}^{q}_i{\bm{E}^{p}_j}^\top, \\
    \hat{\beta}_{i,j}^{q,p} &= \frac{\exp{\left(\beta_{i,j}^{q,p}\right)}}{\sum_{c^p_e\in S^{*}_{c^q_i}}\exp{\left(\beta_{i,e}^{q,p}\right)}}, \\
    \hat{\bm{E}}^{q,*}_i &= {\sum_{c^p_e\in S^{*}_{c^q_i}} \hat{\beta}_{i,j}^{q,p} \bm{E}_j^p}, \\
\end{split}
\end{equation}
where the $*$ could be $V$ and $H$ that represents the horizontal level or vertical level respectively. The embeddings of IPC codes $c^q_i, c^p_j$ are represented by $\bm{E}^q_i, \bm{E}^p_j$ respectively and $p \in [q-1,q,q+1]$. It is worth noticing that we normalize the attention weights, denoted as $\hat{\beta}_{i,j}^{q,p}$, using set $S^{H}_{c^q_i}$ if $R^{i,j}_{q,p}$ is 1. Alternatively, if these two IPC codes at the $q$-th level share the same parent, we normalize using set $S^{V}_{c^q_i}$. To incorporate information from both the vertical and horizontal levels, we concatenate the aggregated information using the function $f(\cdot)$, which is defined as follows,
\begin{equation}
\label{equ:concatenation}
\bm{H}^{MP}_q = f\left([\hat{\bm{E}}^{q,V}, \hat{\bm{E}}^{q,H}]\right),
\end{equation}
where we implement the aggregation function by a 2-layer perceptron neural network.

\textbf{Contextual Representations Generation.} In addition to learning the semantic information across IPC codes at adjacent levels, we concatenate all their parents' embeddings along the taxonomy to learn the global semantic information. The process could be written by,
\begin{equation}
\label{equ:hierar_concatenation}
\bm{H}^P_q =  g_{q}\left(\lVert_{i=1}^{q} \bm{H}^{MP}_{i}\right),
\end{equation}
where we implement the encoding function $g_{q}(\cdot)$ by a 2-layer perceptron neural network for IPC codes at the $q$-th level. The resulting semantic representations denoted as $\bm{H}^P_q\ \in \mathbb{R}^{|C^q|\times 2F}$ capture the contextual information for IPC codes at the $q$-th level.
 
\subsection{Historical Application Patterns Learning}
In general, it has been observed that assignees exhibit certain behavioral patterns when applying for patents. For example, they tend to apply for patent applications in specific fields consistently over a period of time, which is illustrated in \figref{fig:introduction} (a). To capture these historical application behaviors of assignees, we devise a historical application patterns learning module.
For a given patent document $p_k$, we first construct a patent interaction graph for its assignee $u_k$ based on the sequence of historical patents $P^k_{u_k}$. This sequence includes both text and label features for each patent. We then employ the graph convolution network (GCN) \cite{DBLP:conf/iclr/KipfW17} with positional embeddings via a dual message passing mechanism to capture the temporal dependencies from semantics and label two aspects among the patents. Finally, we fuse the aggregated information from all patents in the patent graph for classification. When describing this module, we omit the patent assignee $u_k$ and publication time $t_k$ for simplicity as no ambiguity will be incurred. 

\textbf{Patent Graph Construction.} To build the patent interaction graph $G_{u_k}^{{t_k}}$, we first select the $D$ most recent patent documents applied by assignee $u_k$ prior to patent $p_k$. It ensures that we focus on the latest and most relevant information, allowing us to capture the most up-to-date trends from historical patents. These selected documents serve as the nodes in the graph. In order to capture sequential patterns among historical patents, we employ a sliding window strategy for graph construction. This strategy establishes connections between patent $p_i$ and its previous $s$ patents on the graph $G_{u_k}^{t_k}$ in chronological order. We calculate the weight between patent $p_i$ and $p_j$ ($t_i\leq t_j$) using the following formula,
\begin{equation}
\label{equ:adjacent_mat}
	\bm{A}_{ij} = \begin{cases} 
		\frac{1}{j-i+1} & \text{if } {0\leq j-i<s} ,\\
		0 & \text{others}.
	\end{cases} 
\end{equation}
where $\bm{A} \in \mathbb{R}^{D \times D }$ represents the adjacent matrix of graph $G_{u_k}^{{t_k}}$. It is worth noting that for an assignee's first patent document, there are no preceding historical patents available. As a result, we do not construct the patent graph or propagate message information further for this particular patent.

\textbf{Dual Channels Aggregation.} Within the patent graph $G_{u_k}^{t_k}$, we use two types of features to fully model historical behaviors: textual features $\bm{X}^{c} \in \mathbb{R}^{D \times T}$ and label features $\bm{X}^{s} \in \mathbb{R}^{D \times 2F}$. To be specific, for each patent $p_i$, we utilize the mean function on pre-trained word embeddings $\bm{W}_i$ to represent its textual information. The label features are obtained from one-hot encodings, which can be calculated using the following equation,
\begin{equation}
	\bm{X}^{s}_{i} = \bm{B}^{q}_{i} \bm{W}_B,
\end{equation}
where $\bm{B}^{q}_{i} \in \mathbb{R}^{1 \times |C^q|}$ represents the one-hot vectors of $p_i$'s IPC codes at the $q$-th level (i.e., the specific level for classification), and $\bm{W}_B \in\mathbb{R}^{|C^q|\times 2F}$ is the trainable parameter in transformation operation.

For each patent $p_i$, we employ a graph neural network to aggregate information from its first-order neighbor nodes based on textual and label features. In the first layer, we set the textual features $\bm{X}^{c} $ and label features $\bm{X}^{s}$ as the original features of nodes, denoted as $\bm{H}^{(0)}_{c}$ and $\bm{H}^{(0)}_{s}$ respectively. In order to capture the sequential information explicitly, we take inspiration from the Transformer \cite{DBLP:conf/nips/VaswaniSPUJGKP17} introduce a fixed positional function that utilizes the $\cos{(\cdot)}$ and $\sin{(\cdot)}$ functions for both textual and label features to encode chronological information. It can help us capture the order of the patent documents over time and better utilize the historical information. This can be expressed as follows,
\begin{equation}
\label{equ:position_enc}
\begin{split}
    PE^{*}_{(k, 2i)} &= \sin(\frac{k}{10000^{2i/d_{*}}}),\\
    PE^{*}_{(k, 2i+1)} &= \cos(\frac{k}{10000^{2i/d_{*}}}),\\
    \bm{H}^{(0)}_{*} &= \bm{X}^{*} || PE^{*},
\end{split}
\end{equation}
where $*$ represents either $c$ or $s$, $d_{c}$ and $d_{s}$ correspond to $T$ and $2F$ respectively, $k$ denotes the position of patents in the sequence, and $\bm{H}^{(0)}_{s}$ and $\bm{H}^{(0)}_{c}$ is the initialized embeddings for the graph convolution process. The process of GCN at layer $i$ can be calculated as follows,
\begin{equation}
\label{equ:gcn}
\begin{split}
    \bm{H}^{(i)}_{c} &= ReLU\left(\bm{A}\bm{H}^{(i-1)}_{c}\bm{W}^{(i)}_{c}\right),\\
    \bm{H}^{(i)}_{s} &= ReLU\left(\bm{A}\bm{H}^{(i-1)}_{s}\bm{W}^{(i)}_{s}\right),
\end{split}
\end{equation}
where $\bm{H}^{(i-1)}_{c}, \bm{H}^{(i-1)}_{s}$ is the textual and label embeddings in layer $i-1$, respectively. Additionally, $\bm{W}^{(i)}_{c}, \bm{W}^{(i)}_{s}$ are the trainable weights in the text encoding layer $i$ and label encoding layer $i$.

Finally, to capture the global information from the historical patent sequence, we read out the updated embeddings of all patents at layer $I$. This is achieved through the following equation,
\begin{equation}
\label{equ:readout_all}
    \begin{split}
       \bm{H}_o^s &= Readout(\bm{H}^{(I)}_s), \\
       \bm{H}_o^c &= Readout(\bm{H}^{(I)}_c), \\
       \bm{M}_B &= \left[\bm{H}_o^s,\bm{H}_o^c\right],
    \end{split}
\end{equation}
where $\bm{M}_B\in\mathbb{R}^{2F}$ is the aggregated embedding of patent document $p_k$ based on patent graph $G^{t_k}_{u_k}$. This aggregated embedding is denoted as $\bm{M}^k_B$ in the Prediction module.

\subsection{Prediction Module}
In this section, we focus on making predictions using a combination of textual information and behavioral applications. We first incorporate hierarchical semantic representations from IPC codes into the contextual information of patents, which enriches the patents with meaningful semantic features. Subsequently, we predict the classification probabilities of the patent document $p_k$ with separate information decoders. These decoders take into account both the contextual information of the patent and sequential embeddings of assignees to make their predictions. By considering these factors, our model aims to generate accurate and effective predictions that can be leveraged for various applications.

\textbf{Multi-label Attention Layer.} In multi-label text classification problems, textual information contains abundant characteristics of patent documents and contributes to the classification process \cite{DBLP:conf/aaai/YaoM019, DBLP:conf/ijcai/ShangMXS19, DBLP:conf/cikm/HuangCLCHLZZW19}. In particular, each label represents distinct semantic information that focuses on different aspects of the text. In addition to the embeddings derived from the hierarchical taxonomy, we also capture the static information associated with each IPC code to learn their unique characteristics. Specifically, we utilize trainable parameters $\bm{S}_q$ for IPC codes in the $q$-th level in this work. To leverage these aspects effectively, we employ an attention mechanism that captures the inherent dependencies between patent documents and IPC codes. This attention mechanism allows us to learn patent representations by selectively focusing on relevant information. The process can be summarized as follows,
\begin{equation}
\label{equ:label-attention}
\begin{split}
    \bm{M}_P^k &= softmax\left(\bm{H}^P_q{\bm{V}^k}^{\top}\right)\bm{V}^k, \\
    \bm{M}_{S}^k &= softmax\left(\bm{S}_q{\bm{V}^k}^{\top}\right)\bm{V}^k, \\
    \bm{M}^k_T &= \bm{M}_P^k || \bm{M}_{S}^k
\end{split}
\end{equation}
where $\bm{S}_q, \bm{H}^{P}_q \in\mathbb{R}^{|C^q|\times2F}$ are the static and hierarchical representations for IPC codes, $\bm{V}^k$ denotes the text embeddings of patent $p_k$ and $\bm{M}^k_T \in \mathbb{R}^{|C^q|\times4F}$ denotes the patent text embedding with the incorporation of the semantics of IPC codes.

Then, we combine the aggregated information, including contextual information and behavioral patterns for prediction. The integration is formulated as follows,
\begin{equation}
\label{equ:prediction}
    \hat{\bm{Y}}_k = \sigma\left( g_T\left(\bm{M}^k_T\right) + g_B\left(\bm{M}^k_B\right) \right),
\end{equation}
where $g_T(\cdot), g_B(\cdot)$ refer to two decoders that are implemented using a two-layer perceptron neural network respectively. The output $\hat{\bm{Y}}_k \in \mathbb{R}^{|C^q|}$ represents the prediction probabilities of IPC codes at the $q$-th level, and the sigmoid function $\sigma(\cdot)$ is applied to obtain the final prediction probabilities.

\subsection{Model Optimization} We treat the task of the patent classification problem as a multi-label text classification problem, where each IPC code serves as a label. Specifically, the ground truth for a patent document $p_i$ is denoted as $\bm{Y}_{i}\in\{0,1\}^{|C^q|}$, where we only predict the IPC codes at the third level, following the approaches presented in \cite{DBLP:conf/aaai/Tang0XPWC20, DBLP:conf/cikm/HuangCLCHLZZW19}. In this representation, $\bm{Y}_{i,j}$ equals $1$ if the label $c^q_j$ is assigned to patent $p_i$. During the training process, we convert the multi-label text classification task into several binary classification problems, by creating a binary classification sub-problem for each label independently. We optimize the model parameters by minimizing cross-entropy loss, which can be expressed as follows,

\begin{equation}
\label{equ:optimization}
    L = -\sum_{p_i \in \mathbb{P}}\sum_{c_j^q \in {C}^q} \bm{Y}_{i,j}\log(\hat{\bm{Y}}_{i,j}) + (1-\bm{Y}_{i,j})\log(1-\hat{\bm{Y}}_{i,j}),
\end{equation}
where $\mathbb{P}$ represents the set of patent documents and $C^q$ denotes the set of IPC codes at the $q$ level.

\begin{algorithm}[!htbp]
\SetKwComment{Comment}{/* }{ */}
\SetKwInOut{Input}{Input}
\SetKwInOut{Output}{Output}
\caption{Training process of our model}
\label{alg:training_process}
\Input{Collection of patent documents $\mathbb{P}=\{p_1,p_2,\cdots, p_k\}$, where patent $p_i$ is represented as $p_i=(W_i,u_i,t_i,\bm{Y}_i)$, maximum number of training epochs $MaxEpoch$\;}
\Output{The model parameters $\Theta$ after training\;}
Initialize the parameters in our model with random weights $\Theta$\ and  set $Epochs \gets 1$\;
\While{not converge and $Epochs \leq MaxEpoch$}{
    \For{batch $B_b \in \mathcal{B}$ }{
        $\{\bm{V}^{k}\}$ $\gets$ Encode the textual representations of patent documents in $B_b$ using a pre-trained model and the Bi-LSTM mechanism with the description text of patents, denoted as $\{W_k\}$ as inputs, where $W_k=\left[w^k_1, w^k_2, \cdots, w^k_N\right]$\;
        $\{\bm{H}^P_q\}$ $\gets$ Encode the contextual embeddings for IPC codes at the $q$-th level in the taxonomy via \equref{equ:new hierarchy_horizon}, \myref{equ:concatenation} and \myref{equ:hierar_concatenation} with trainable parameters of IPC codes $\{\bm{E}^1,\cdots,\bm{E}^L\}$, and the taxonomy structure $\mathcal{R}=\{\bm{R}^{1,2},\cdots,\bm{R}^{L-1, L}\}$ as inputs\; 
        $\{G^{t_k}_{u_k}\}$ $\gets$ Build the patent interaction graph for assignee $u_k$ using a sliding window strategy via \equref{equ:adjacent_mat} with its recent $D$ patent documents as inputs\;
        $\{\bm{M}_{B}^{k}\}$ $\gets$ Encode the behavior patterns of assignee using a dual channels aggregation mechanism $u_k$ via \equref{equ:gcn} and \myref{equ:readout_all} with textual features $\bm{X}^c$, and label features $\bm{X}^s$ from its historical patents, which are enhanced with positional embeddings generated by \equref{equ:position_enc} as inputs\;
        $\{\hat{\bm{Y}}_k\}$ $\gets$ Predict the probabilities of IPC codes at the $q$-th level via \equref{equ:label-attention} and \myref{equ:prediction} with textual information $\bm{V}^k$, contextual embeddings $\bm{H}^P_q$ and static information $\bm{S}_q$ of IPC codes, and behaviors representations $\bm{M}^k_B$ for assignee $u_k$ as inputs\;

        Train our model via \equref{equ:optimization} with $\{\hat{\bm{Y}}_k\}$ and $\{\bm{Y}_k\}$ as inputs\;

    }
    $Epochs \gets Epochs + 1$\;
}
\end{algorithm}

%% file: section-5-experiments.tex
\section{Experiments}
\label{section-4}
In this section, we conduct extensive experiments on real-world datasets to evaluate the model performance. 

\subsection{Datasets Descriptions}
In this work, we use two patent datasets to evaluate the proposed model. The statistical information of these datasets is presented in \tabref{tab:datasets_information}, where \#C denotes the number of assignees. \#train, \#validate, and \#test correspond to the number of patent documents in training, validation, and testing sets, respectively. Furthermore, we provide \#IPC codes, which represent the total number of IPC codes in each dataset, with L1, L2, and L3 denoting the amount of IPC codes at the first level, second level, and third level, respectively. We provide a detailed overview of these two datasets as follows.

\begin{itemize}
    \item USPTO-200K is a collection of patent documents retrieved from the USPTO\footnote{\url{https://www.uspto.gov/}}. Due to the absence of certain metadata, such as assignee information or timestamps, in some patent documents, we apply a filtering process to select a subset of the complete collection. To be specific, we specifically choose patent documents from the period 2010 to 2019 that contain essential information including assignee details, text descriptions, and IPC codes. This selection process ensures that the dataset we work with includes relevant patents with comprehensive information for our analysis and research. For data partition, we chose the patent documents in 2010-2017, 2018, and 2019 as training datasets, validation datasets, and testing datasets. Furthermore, we remove the patents whose assignees have no patents in training, validation, or testing. 
    \item CNPTD-200K records the patent documents of technology companies in China from 2010 to 2019 for experiments. Similar to the USPTO-200K, we select patent documents from the years 2010-2017, 2018, and 2019 as our training, validation, and testing datasets, respectively. In addition, we exclude patent documents where the assignees have no patents in the training, validation, or testing sets.

\end{itemize}

\begin{table}[!htbp]
\centering
\caption{Statistics of the two datasets.}
\label{tab:datasets_information}
\resizebox{0.6\columnwidth}{!}
{
\setlength{\tabcolsep}{1.0mm}
{
\begin{tabular}{c|c|ccc|ccc}
\hline
\multirow{2}{*}{Datasets} & \multirow{2}{*}{\#C} & \multicolumn{3}{c|}{\#patents} & \multicolumn{3}{c}{\#IPC codes} \\ \cline{3-8} 
                          &                        & \#train        & \#validate       & \#test      & L1       & L2        & L3       \\ \hline
USPTO-200K                & 513                    & 152,349   & 24,335   & 23,675  & 8        & 125       & 622      \\
CNPTD-200K                & 1,443                  & 120,130   & 40,128   & 39,807  & 8        & 124       & 608      \\ \hline
\end{tabular}
}
}
\end{table}


\subsection{Compared Methods}
We compare our approach with six baselines. Word2Vec \cite{DBLP:journals/corr/abs-1301-3781}, Bi-LSTM \cite{DBLP:journals/corr/Graves13}, and DeepPatent \cite{DBLP:journals/scientometrics/LiHCH18} capture contextual information solely from text descriptions for classification. AttentionXML \cite{DBLP:conf/nips/YouZWDMZ19} is one of the representative methods in multi-label text classification tasks. HARNN \cite{DBLP:conf/cikm/HuangCLCHLZZW19} belongs to the models in hierarchical multi-label text classification problems. At last, A-GCN+A-NLSOA \cite{DBLP:conf/aaai/Tang0XPWC20} represents one of the recent state-of-the-art methods in patent categorization. Details of baselines are introduced as follows.

\begin{itemize}
	\item  \textbf{Word2Vec} \cite{DBLP:journals/corr/abs-1301-3781} learns word embeddings with the assumption that words with similar meanings tend to be used in similar contexts within a corpus of texts. We further apply the MEAN function to capture the textual dependencies across words, with a sigmoid function and binary cross-entropy loss for training. The baselines listed below all apply Word2Vec to obtain word embeddings as an initial step.

    \item  \textbf{Bi-LSTM} \cite{DBLP:journals/corr/Graves13} is a type of recurrent neural network architecture that utilizes bidirectional Long Short-Term Memory (LSTM) units to learn the contextual embedding within the text sequence. In this work, we then apply a sigmoid function and binary cross-entropy loss for classification. 

    \item  \textbf{DeepPatent} \cite{DBLP:journals/scientometrics/LiHCH18} leverages convolution neural networks and the max-pooling module on patent documents to learn both local and global dependencies within the text sequence and then uses fully connected layers to predict the classification probabilities. 

    \item  \textbf{AttentionXML} \cite{DBLP:conf/nips/YouZWDMZ19} applies an attention mechanism on text descriptions to capture the most relevant information to each label. Additionally, it incorporates a probabilistic label tree to efficiently handle large label sets, improving overall efficiency.

    \item  \textbf{HARNN} \cite{DBLP:conf/cikm/HuangCLCHLZZW19} designs a recurrent-based neural network to learn the hierarchical semantic information in a top-down manner that incorporates an attention mechanism in each layer. This attention mechanism captures relationships between labels at each level and the information contained in patent texts, enabling HARNN to effectively analyze the hierarchical structure of IPC codes.

    \item \textbf{A-GCN+A-NLSOA} \cite{DBLP:conf/aaai/Tang0XPWC20} employs
    graph neural networks and recurrent neural networks to both learn sequential and high-order dependencies between words in the text corpus, with an attention mechanism to capture the relationships between text descriptions and IPC codes for classification.

    \item \textbf{HmcNet} \cite{huang2022hmcnet} integrates implicit class hierarchy constraints to capture class dependencies in the recurrent-based learning process and explicit class hierarchy information to generate coherent predictions. 
\end{itemize}

\begin{table*}[!htbp]
\centering
\caption{Performance of different methods on USPTO-200K and CNPTD-200K datasets. ${\star}$ indicates the improvement is statistically significant with a paired t-test (i.e., $p$ \textless 0.05).}
\label{tab:performance_comparison}
\resizebox{1\columnwidth}{!}
{
\begin{tabular}{c|c|ccc|ccc|ccc}
\hline
\multirow{2}{*}{Datasets} & \multirow{2}{*}{Methods} & \multicolumn{3}{c|}{K=1}                                                                                                             & \multicolumn{3}{c|}{K=3}                                                                                                             & \multicolumn{3}{c}{K=5}                                                                                                              \\ \cline{3-11} 
                          &                          & Precision                                  & Recall                                     & NDCG                                       & Precision                                  & Recall                                     & NDCG                                       & Precision                                  & Recall                                     & NDCG                                       \\ \hline
\multirow{9}{*}{USPTO-200K}                & Word2Vec                 & 0.7706                                     & 0.4817                                     & 0.7706                                     & 0.4536                                     & 0.7336                                     & 0.7482                                     & 0.3142                                     & 0.8111                                     & 0.7742                                     \\
                          & Bi-LSTM                  & 0.7966                                     & 0.4994                                     & 0.7966                                     & 0.4607                                     & 0.7457                                     & 0.7651                                     & 0.3175                                     & 0.8197                                     & 0.7893                                     \\
                          & DeepPatent               & 0.7910                                     & 0.4959                                     & 0.7910                                     & 0.4560                                     & 0.7405                                     & 0.7590                                     & 0.3152                                     & 0.8156                                     & 0.7840                                     \\
                          & AttentionXML             & 0.8123                                     & 0.5080                                     & 0.8123                                     & 0.4727                                     & 0.7633                                     & 0.7817                                     & 0.3250                                     & 0.8390                                     & 0.8059                                     \\
                          & HARNN                    & 0.8155                                     & 0.5100                                     & 0.8155                                     & 0.4741                                     & 0.7632                                     & 0.7838                                     & 0.3272                                     & 0.8384                                     & 0.8084                                     \\
                          & A-GCN+A-NLSOA            & 0.8171                                     & 0.5114                                     & 0.8171                                     & 0.4743                                     & 0.7632                                     & 0.7848                                     & 0.3272                                     & 0.8381                                     & 0.8092                                     \\
                          & HmcNet                   & 0.8179                                     & 0.5118                                     & 0.8179                                     & 0.4762                                     & 0.7661                                     & 0.7871                                     & 0.3274                                     & 0.8395                                     & 0.8105                                     \\
                          & Ours                     & \textbf{0.8341}$^{\star}$ & \textbf{0.5226}$^{\star}$ & \textbf{0.8341}$^{\star}$ & \textbf{0.4856}$^{\star}$ & \textbf{0.7794}$^{\star}$ & \textbf{0.8017}$^{\star}$ & \textbf{0.3342}$^{\star}$ & \textbf{0.8537}$^{\star}$ & \textbf{0.8255}$^{\star}$ \\
                          & Improvements             & 1.98\%                                     & 2.11\%                                     & 1.98\%                                     & 1.97\%                                     & 1.74\%                                     & 1.85\%                                     & 2.08\%                                     & 1.69\%                                     & 1.85\%                                     \\ \hline
\multirow{9}{*}{CNPTD-200K}                & Word2Vec                 & 0.5945                                     & 0.5005                                     & 0.5945                                     & 0.3025                                     & 0.7146                                     & 0.661                                      & 0.2033                                     & 0.7857                                     & 0.6922                                     \\
                          & Bi-LSTM                  & 0.6382                                     & 0.5405                                     & 0.6382                                     & 0.3134                                     & 0.7437                                     & 0.6956                                     & 0.2082                                     & 0.8084                                     & 0.7240                                     \\
                          & DeepPatent               & 0.6431                                     & 0.5444                                     & 0.6431                                     & 0.3159                                     & 0.7495                                     & 0.7007                                     & 0.2102                                     & 0.8161                                     & 0.7300                                     \\
                          & AttentionXML             & 0.6563                                     & 0.5566                                     & 0.6563                                     & 0.3286                                     & 0.7843                                     & 0.7230                                     & 0.2174                                     & 0.8487                                     & 0.7510                                     \\
                          & HARNN                    & 0.6537                                     & 0.5532                                     & 0.6537                                     & 0.3206                                     & 0.7599                                     & 0.7114                                     & 0.2122                                     & 0.8227                                     & 0.7390                                     \\
                          & A-GCN+A-NLSOA            & 0.6695                                     & 0.5660                                     & 0.6695                                     & 0.3376                                     & 0.7961                                     & 0.7403                                     & 0.2237                                     & 0.8617                                     & 0.7693                                     \\
                          & HmcNet                   & 0.6790                                      & 0.5743                                     & 0.6790                                      & 0.3393                                     & 0.7999                                     & 0.7462                                     & 0.2241                                     & 0.8635                                     & 0.7742                                     \\
                          & Ours                     & \textbf{0.6848}$^{\star}$ & \textbf{0.5795}$^{\star}$ & \textbf{0.6848}$^{\star}$ & \textbf{0.3428}$^{\star}$ & \textbf{0.8084}$^{\star}$ & \textbf{0.7540}$^{\star}$ & \textbf{0.2257}$^{\star}$ & \textbf{0.8708}$^{\star}$ & \textbf{0.7814}$^{\star}$ \\
                          & Improvements             & 0.85\%                                     & 0.91\%                                     & 0.85\%                                     & 1.03\%                                     & 1.06\%                                     & 1.05\%                                     & 0.71\%                                     & 0.85\%                                     & 0.93\%                                     \\ \hline
\end{tabular}

}

\end{table*}

\subsection{Evaluation Metrics}
Three metrics are adopted to comprehensively evaluate the performance of different models, including Precision, Recall, and Normalized Discounted Cumulative Gain (NDCG). 

Precision quantifies the accuracy of recognizing relevant elements in comparison to irrelevant ones. For patent $p_i$, Precision is calculated by
\begin{equation}
    \notag
    \mathrm{Precision@K}(p_i) = \frac{|\hat{S}_i \cap S_i|}{|\hat{S}_i|},
\end{equation}
where $\hat{S}_i$ and $S_i$ are the predicted top-K labels and the real labels of $p_i$, respectively. $|S|$ is the size of the set $S$. We use the average Precision of all patents as a metric.

Recall assesses the model's capability to identify and select all relevant elements. For patent $p_i$, Recall is calculated by
\begin{equation}
    \notag
    \mathrm{Recall@K}(p_i) = \frac{|\hat{S}_i \cap S_i|}{|S_i|},
\end{equation}
where the meaning of $\hat{S}_i$, $S_i$, and $|S|$ remains consistent with the definition in the Precision metric, and we calculate the average Recall across all patents as a measure.

NDCG evaluates the quality of rankings by taking into account the order of all labels. For patent $p_i$, NDCG is computed by
\begin{equation}
    \notag
    \mathrm{NDCG@K}(p_i) = \frac{\sum_{k = 1}^{K}{\frac{\delta(\hat{S}_i^k, S_i)}{\log_2(k + 1)}}}{\sum_{k = 1}^{\min(K, |S_i|)}{\frac{1}{\log_2\left(k + 1\right)}}},
\end{equation}
where $\hat{S}_i^k$ represents the $k$-th predicted label of $p_i$, and $\delta\left(v,S\right)$ equals 1 if element $v$ is in set $S$, otherwise 0. We calculate the average NDCG of all patents as a metric.

\subsection{Experimental Settings}
In this work, we use the ``Summary'' section as the text description for each patent document, which describes the brief information about patents. Besides, We treat the IPC codes at the L3 level as truth labels, which follow the existing works \cite{DBLP:conf/aaai/Tang0XPWC20, DBLP:conf/cikm/HuangCLCHLZZW19}. We select all the text descriptions in training datasets and remove words appearing less than 5 times to build the text corpus. Then we use Gensim \textit{Word2Vec} tool \footnote{https://radimrehurek.com/gensim/models/word2vec.html} to train the 100-dimensional word embedding for each word in the text corpus. For those words not in the text corpus, they are initialized with the MASK embedding, which is set randomly. All the word embeddings are trainable in the models. In our experiments, we choose the first 100 words in the ``Summary'' of each document for training, validation, and testing, followed by \cite{DBLP:conf/aaai/Tang0XPWC20}. We use the grid search to identify the number of $D$ and $s$ in USPTO-200K and CNPTD-200K, i.e., (50, 10), (40,10) respectively. In the training process, we set the learning rate to 0.0001 and use Adam \cite{kingma2014adam} as the optimizer. To prevent over-fitting, we use dropout \cite{DBLP:journals/jmlr/SrivastavaHKSS14} with a dropout rate of 0.5 in our model. For all the methods, we search the hidden dimension in [64, 128, 256, 512] and apply grid search to select the best dropout and learning rate. We set the training epoch 300 on all experiments and use an early stopping strategy with a patience of 10. Lastly, we choose the model that achieves the best performance on the validation for testing. The codes and datasets are available at https://github.com/Hope-Rita/PatCLS. 

\subsection{Performance Comparison}
\label{subsec:performance_all_patents}
The comparisons of all the methods on top-K performance are shown in \tabref{tab:performance_comparison}. We can observe that our model outperforms existing methods in all metrics, demonstrating its effectiveness in patent classification problems. Moreover, we further summarize several conclusions as follows.

Firstly, Bi-LSTM and DeepPatent demonstrate significantly better performance compared to Word2Vec, suggesting that deep learning algorithms such as RNN \cite{DBLP:journals/neco/HochreiterS97} and CNN \cite{DBLP:conf/nips/CunBDHHHJ89} can effectively capture the contextual information in text descriptions through their inherent ability to model the sequential semantics.

Secondly, AttentionXML outperforms Bi-LSTM and DeepPatent because AttentionXML could capture the dependencies between label and text information. By leveraging attention mechanisms, AttentionXML can effectively learn the relevant parts of texts to make better predictions.

Thirdly, HARNN exhibits superiority over AttentionXML in the USPTO-200K, indicating the benefits of utilizing hierarchical information of IPC codes for patent classification tasks. However, HARNN fails to outperform AttentionXML in CNPTD-200K, which suggests that HARNN may struggle to effectively learn and utilize the hierarchical structure information in CNPTD-200K.

Fourthly, A-GCN+A-NLSOA achieves better performance than the above baselines due to its capability of combining intricate text relationships, such as non-local word dependencies and fine-grained semantic information. HmcNet obtains the best performance in all baselines since it captures the correlations of classification codes at the adjacent levels and integrates the hierarchy constraints into the path selection process. In addition, we observe that the NDCG tends to decrease when K is small in certain models. This is because the increase in correctly classified labels is smaller compared to the increase in K, resulting in a negative correlation between NDCG and K. However, the NDCG exhibits a positive trend as K increases. It suggests that the NDCG would improve when a larger number of items are considered (i.e., with a sufficiently large K value), which leads to better evaluation.

Finally, our model outperforms the existing methods due to the following reasons. Firstly, it effectively captures the semantic relationships among IPC codes by leveraging the taxonomy structure adaptively, which allows us to learn informative related parts of texts to each label. Secondly, it discovers the sequential application behaviors of assignees, providing valuable information for accurate classification of the current patent document.

\subsection{Effects of Message Passing in the IPC Taxonomy}

\begin{table}[!htbp]
\centering
\caption{Effects of convolution in a hierarchical taxonomy on USPTO-200K and CNPTD-200K datasets. }
\label{tab:hierarchy_comparison}
\resizebox{0.85\columnwidth}{!}
{
\begin{tabular}{c|c|cc|cc|cc}
\hline
\multirow{2}{*}{Datasets}   & \multirow{2}{*}{Methods} & \multicolumn{2}{c|}{K=1}          & \multicolumn{2}{c|}{K=3}          & \multicolumn{2}{c}{K=5}           \\ \cline{3-8} 
                            &                          & Recall          & NDCG            & Recall          & NDCG            & Recall          & NDCG            \\ \hline
\multirow{5}{*}{USPTO-200K} & PSE                  & 0.5111          & 0.8182          & 0.7644          & 0.7856          & 0.8375          & 0.8090          \\
                            & w/ fixed Aggregation        & 0.5119          & 0.8186          & 0.7657          & 0.7864          & 0.8397          & 0.8104          \\
                            & w/ adaptive v-inf       & 0.5126          & 0.8188          & 0.7680          & 0.7885          & 0.8421          & 0.8126          \\
                            & w/ adaptive h-inf         & 0.5129          & 0.8195          & 0.7665          & 0.7877          & 0.8419          & 0.8122          \\
                            & PSE+ICL                     & \textbf{0.5138} & \textbf{0.8207} & \textbf{0.7681} & \textbf{0.7898} & \textbf{0.8433} & \textbf{0.8141} \\ \hline
\multirow{5}{*}{CNPTD-200K} & PSE                  & 0.5608          & 0.6627          & 0.7779          & 0.7267          & 0.8419          & 0.7548          \\
                            & w/ fixed Aggregation        & 0.5710          & 0.6745          & 0.7971          & 0.7433          & 0.8613          & 0.7716          \\
                            & w/ adaptive v-inf       & 0.5720          & 0.6763          & 0.7973          & 0.7439          & 0.8593          & 0.7711          \\
                            & w/ adaptive h-inf         & 0.5733          & 0.6773          & 0.7992          & 0.7454          & 0.8631          & 0.7743          \\
                            & PSE+ICL                     & \textbf{0.5748} & \textbf{0.6800} & \textbf{0.8002} & \textbf{0.7470} & \textbf{0.8636} & \textbf{0.7749} \\ \hline 
\end{tabular}
}
\end{table}
As our work adaptively learns the structure information among IPC codes in the taxonomy, it can capture the inherent dependencies across IPC codes at different levels, which helps our model incorporate semantic relations on IPC codes for prediction. Hence, we validate the ability to learn semantic relationships in the IPC codes Correlations Learning (ICL) module by comparing them with the other three variants models. To reduce the external effects from other aspects, we use the Patent Text Embedding (PSE) module, which just utilizes textual information that integrates the static learnable representations of IPC codes for prediction as our basic model. Specifically, we use three variant models based on PSE for experiments. Firstly, we aggregate the semantic information among IPC codes along the fixed hierarchical taxonomy via \equref{equ:hierar_concatenation}, namely w/ fixed Aggregation. Secondly, we learn the vertical dependencies among IPC codes at the adjacent levels or horizontal correlations across IPC codes at the same levels according to the belongingness relationships adaptively via \equref{equ:new hierarchy_horizon}, namely w/ adaptive v-inf and w/ adaptive h-inf. At last, we adaptively propagate message information across IPC codes in both horizontal and vertical perspectives via the attention mechanism in \equref{equ:new hierarchy_horizon} and fuse the hierarchical information along the taxonomy with \equref{equ:hierar_concatenation} to learn the complex relationships, namely PSE+ICL. We conduct five runs for all experiments and present the average results on Recall and NDCG when the values of K are set to 1, 3, and 5 at \tabref{tab:hierarchy_comparison}. We obtain similar observations on Precision and thus do not plot them due to space limitations.

Based on the results presented in \tabref{tab:hierarchy_comparison}, several observations can be made. Firstly, we find that by aggregating information from the fixed correlation matrix, our model effectively captures semantic relationships across IPC codes. This indicates that learning the semantic dependencies among IPC codes is advantageous for classification tasks. Secondly, as shown in 
 \figref{fig:introduction} (b), we observe diverse relationships between IPC codes in different datasets. Therefore, incorporating adaptive correlations between IPC codes enables our model to enhance the comprehensive semantic relationships within the hierarchical taxonomy. Lastly, while existing approaches primarily focus on capturing either vertical or horizontal correlations in the hierarchy, our model successfully learns both vertical and horizontal semantics between IPC codes, leading to improved performance.

\subsection{Effects of Capturing Historical Behavior Patterns}

\subsubsection{High-order Complex Behaviors Modelling}

\begin{figure}[!htbp]
    \centering
    \includegraphics[width=0.88\columnwidth]{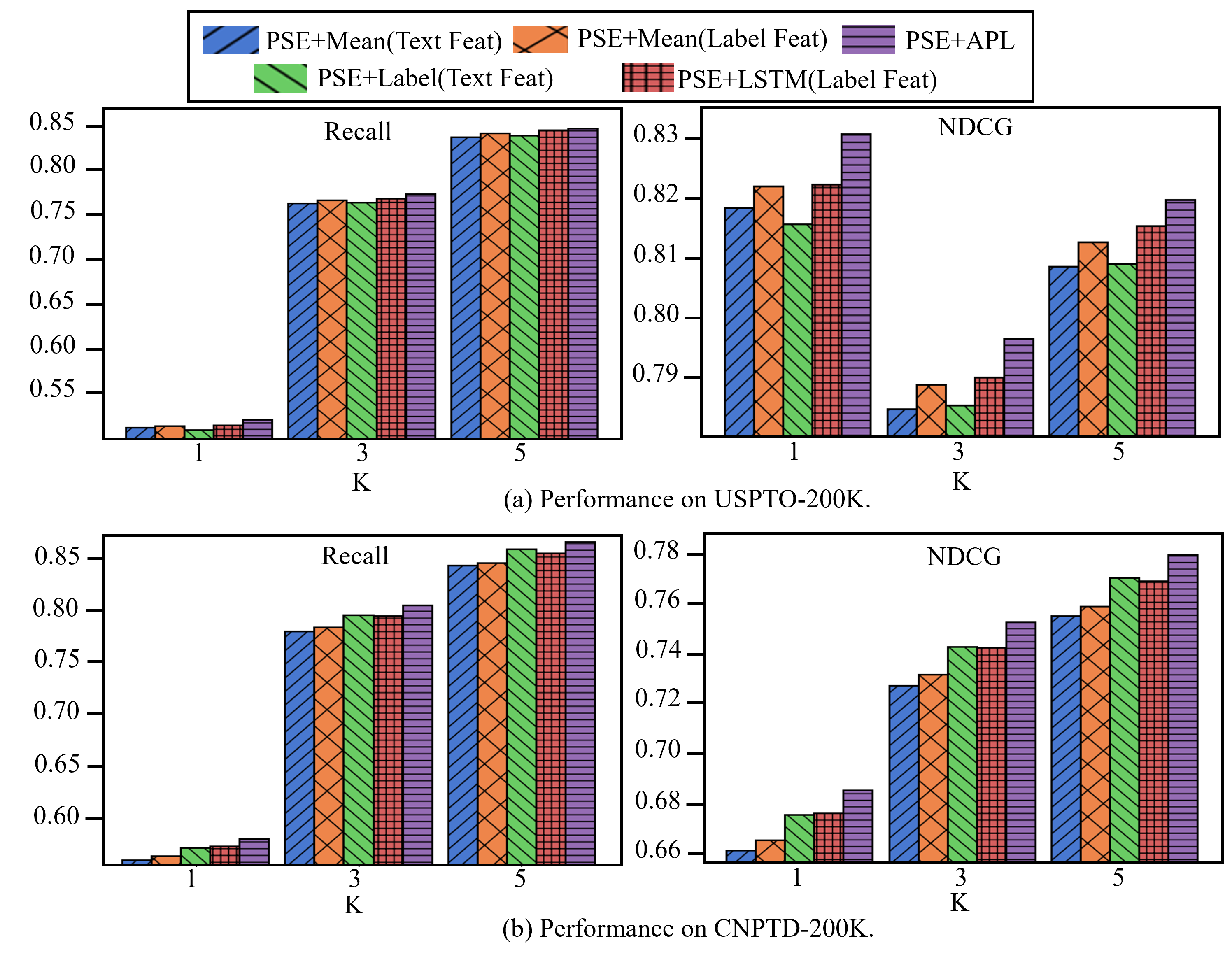}
    \caption{Performance on capturing high-order complex behavior patterns.}
    \label{fig:Effect of Sequential model}
\end{figure}

To model the sequential publication patterns of assignees, we apply a graph neural network with a dual aggregation mechanism based on the constructed individual patent graph for each assignee when classifying new patents. To validate the ability to learn sequential patterns of assignees, we apply four other methods to learn historical sequential information, which fuse text information for classification. In particular, ``Text Feat'' and ``Label Feat'' represent the type of input features in the model, e.g., textual features and label features respectively.
PSE+Mean aggregates the textual information from the Patent Text Embedding module (referred to as PSE) with a mean function, while PSE+LSTM means that we apply a type of recurrent neural network, i.e., LSTM mechanism to capture the sequential relationships. We combine the Patent Text Embedding and Historical Application Patterns Learning (referred to as APL) modules for comparison, namely PSE+APL.

Upon analyzing the results depicted in \figref{fig:Effect of Sequential model}, it is evident that PSE+LSTM consistently outperforms PSE+Mean, indicating the presence of sequential dependencies within historical patents. Nevertheless, these models solely capture sequential information and overlook the underlying high-order structural preferences. To address this limitation, we adopt a sliding window strategy to construct the patent graph. This approach guarantees the preservation of sequential information along publication times and enables the graph neural network to capture high-order information for patents applied within a specific period directly.

\subsubsection{Analysis of Varied Lengths on Historical Patents}
\begin{figure}[!htbp]
    \centering
    \includegraphics[width=0.9\columnwidth]{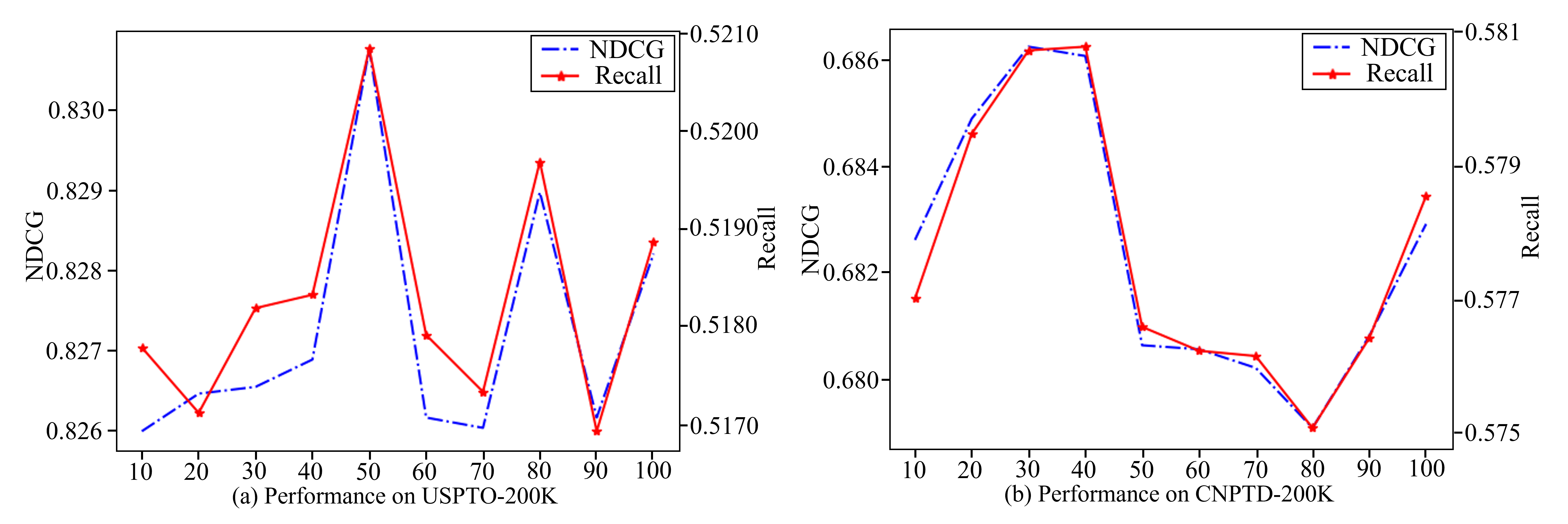}
    \caption{Performance of methods with varied lengths of historical patents for capturing assignees' historical patterns.}
    \label{fig: Data ratio}
\end{figure}

In this work, we use the latest $D$ patent documents to build the patent interaction graph for each assignee to obtain recent behavior publications or trends. In particular, apart from the first patent applied by each assignee, all the remained ones could utilize the sequence of historical patents to capture historical behavior patterns. We further evaluate the effectiveness of varied lengths on historical patents in capturing behavior patterns for assignees. Specifically, we vary the number of historical patents $D$ from 10 to 100, using a sliding window of size 10. Then we predict the probabilities of IPC codes with textual information and historical behavior representations of assignees. The results are shown in \figref{fig: Data ratio}.

From \figref{fig: Data ratio}, it indicates that our model could achieve better performance when the number of historical patents is set at a proper range. The result would be worse when the number is too small or too large, which may be because of the insufficient temporal patterns captured by our model. Hence, choosing a suitable number of historical patents for capturing temporal behavior patterns on different datasets is also essential for satisfactory prediction performance.

\subsubsection{Effects of Capturing Informative Assignees' Preferences}
In our work, we discover that learning the publication patterns of assignees can greatly contribute to patent classification. In this part, we aim to examine the impact of two features within the patent graph and the influence of position encoding for patents. To isolate the effects from other aspects, we solely rely on textual information and historical patterns for prediction. Specifically, we exclude the textual and label information when aggregating messages in the historical application patterns learning module, referred to as w/o Text and w/o Label respectively. Additionally, we eliminate the positional encoding for patents, denoted as w/o PE. We show the performance of Recall and NDCG metrics with K values set at 1, 3, and 5 in \tabref{tab:Effect of Text and Label}.

\begin{table}[!htbp]
\centering
\caption{Performance on capturing informative historical patterns on USPTO-200K and CNPTD-200K datasets.}
\label{tab:Effect of Text and Label}
\resizebox{0.8\columnwidth}{!}
{
\begin{tabular}{c|c|cc|cc|cc}
\hline
\multirow{2}{*}{Datasets}   & \multirow{2}{*}{Methods} & \multicolumn{2}{c|}{K=1}          & \multicolumn{2}{c|}{K=3}          & \multicolumn{2}{c}{K=5}           \\ \cline{3-8} 
                            &                          & Recall          & NDCG            & Recall          & NDCG            & Recall          & NDCG            \\ \hline
\multirow{4}{*}{USPTO-200K} & w/o PE                   & 0.5183          & 0.8269          & 0.7720          & 0.7949          & 0.8451          & 0.8163          \\
                            & w/o Label                & 0.5156          & 0.8237          & 0.7705          & 0.7921          & 0.8432          & 0.8162          \\
                            & w/o Text                 & 0.5200          & 0.8292          & 0.7723          & 0.7951          & 0.8455          & 0.8169          \\
                            & Ours                     & \textbf{0.5208} & \textbf{0.8308} & \textbf{0.7733} & \textbf{0.7964} & \textbf{0.8467} & \textbf{0.8197} \\ \hline
\multirow{4}{*}{CNPTD-200K} & w/o PE                   & 0.5774          & 0.6823          & 0.8033          & 0.7510          & 0.8632          & 0.7779          \\
                            & w/o Label                & 0.5696          & 0.6733          & 0.7962          & 0.7419          & 0.8598          & 0.7700          \\
                            & w/o Text                 & 0.5780          & 0.6834          & 0.8038          & 0.7508          & \textbf{0.8678} & 0.7790          \\
                            & Ours                     & \textbf{0.5808} & \textbf{0.6861} & \textbf{0.8061} & \textbf{0.7531} & 0.8675          & \textbf{0.7802} \\ \hline
\end{tabular}
}
\end{table}

Based on the findings presented in \tabref{tab:Effect of Text and Label}, it can be concluded that our model achieves the best performance when combining the above information. Specifically, textual information provides semantic relationships within the sequence of assignees' historical patent documents while label information captures the research preferences of assignees. Furthermore, the inclusion of position encoding enhances sequential order information among historical patents. By leveraging all of this information simultaneously, we are able to effectively capture the comprehensive preferences of assignees for classifying patent documents.

\subsection{Performance on Patent Classification at Higher Levels}

\begin{figure}[!htbp]
    \centering
    \includegraphics[width=0.80\columnwidth]{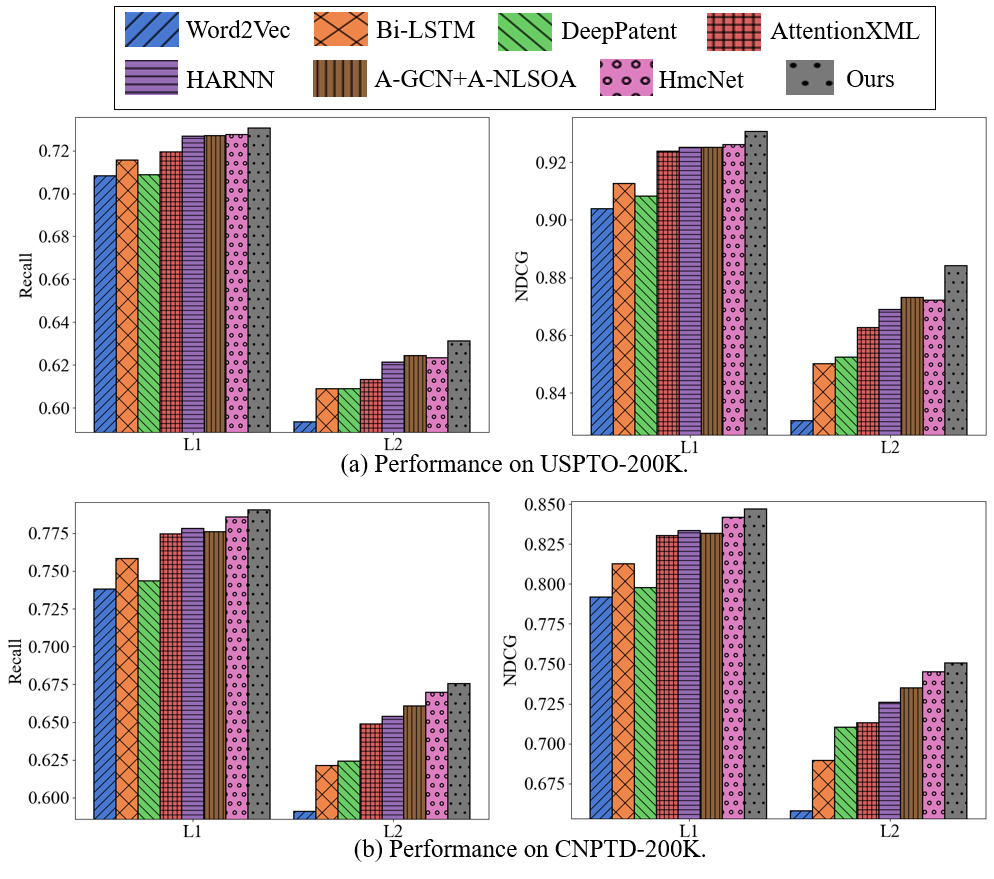}
    \caption{Performance on patent classification at different levels.}
    \label{fig:different levels}
\end{figure}

Additionally, apart from predicting classification codes at the lowest level (i.e., the L3 level), we also validate the performance when predicting classification codes at higher levels, specifically at the first level (i.e., the L1 level) and second level (i.e., the L2 level). Performance of all the methods on the Recall and NDCG metrics is shown in \figref{fig:different levels} when K is set as 1. Due to space constraints, the Precision metric and the performance for K values of 3 and 5 exhibit similar trends.

From \figref{fig:different levels}, we observe that the performance improves when making predictions at higher levels. This can be attributed to the smaller number of classification codes at higher levels, which makes the task relatively easier. Furthermore, our model demonstrates superior performance compared to the baselines, primarily due to its ability to capture fine-grained and bi-directional semantic correlations among classification codes along the taxonomy structure while the baselines only capture top-down vertical relationships across classification codes. Additionally, it effectively learns the sequential historical application behaviors information of each assignee, further contributing to its enhanced performance.

\subsection{Effects of Different Components for Classification}
\label{subsec:alation std}

\begin{table}[!htbp]
\centering
\caption{Performance of different variants on USPTO-200K and CNPTD-200K datasets.}
\label{tab:Ablation_Study}
\resizebox{1\columnwidth}{!}
{
\begin{tabular}{c|c|ccc|ccc|ccc}
\hline
\multirow{2}{*}{Datasets}   & \multirow{2}{*}{Methods} & \multicolumn{3}{c|}{K=1}                            & \multicolumn{3}{c|}{K=3}                            & \multicolumn{3}{c}{K=5}                             \\ \cline{3-11} 
                            &                          & Precision       & Recall          & NDCG            & Precision       & Recall          & NDCG            & Precision       & Recall          & NDCG            \\ \hline
\multirow{4}{*}{USPTO-200K} & PSE                      & 0.8182          & 0.5111          & 0.8182          & 0.4749          & 0.7644          & 0.7856          & 0.3269          & 0.8375          & 0.809           \\
                            & PSE+ICL                  & 0.8207          & 0.5138          & 0.8207          & 0.4781          & 0.7681          & 0.7898          & 0.3295          & 0.8433          & 0.8141          \\
                            & PSE+APL                  & 0.8308          & 0.5208          & 0.8308          & 0.4810          & 0.7733          & 0.7964          & 0.3307          & 0.8467          & 0.8197          \\

                            & Ours                     & \textbf{0.8328} & \textbf{0.5210} & \textbf{0.8328} & \textbf{0.4837} & \textbf{0.7766} & \textbf{0.7992} & \textbf{0.3332} & \textbf{0.8515} & \textbf{0.8233} \\ \hline
\multirow{4}{*}{CNPTD-200K} & PSE                      & 0.6627          & 0.5608          & 0.6627          & 0.3293          & 0.7779          & 0.7267          & 0.2180          & 0.8419          & 0.7548          \\
                            & PSE+ICL                  & 0.6800          & 0.5748          & 0.6800          & 0.3399          & 0.8002          & 0.7470          & 0.2243          & 0.8636          & 0.7749          \\
                            & PSE+APL                  & 0.6861          & 0.5808          & 0.6861          & 0.3420          & 0.8061          & 0.7531          & 0.2251          & 0.8675          & 0.7802          \\
                            & Ours                     & \textbf{0.6873} & \textbf{0.5817} & \textbf{0.6873} & \textbf{0.3444} & \textbf{0.8102} & \textbf{0.7564} & \textbf{0.2267} & \textbf{0.8719} & \textbf{0.7835} \\ \hline
\end{tabular}
}
\end{table}

We validate the effectiveness of the Patent Text Embedding (PSE), the IPC codes Correlations Learning (ICL), and the historical Application Patterns Learning (APL) by removing these three components and then comparing their performance with our model. 
Specifically, PSE just utilizes contextual information for prediction. PSE+ICL adds semantic dependencies among IPC codes based on the PSE model. PSE+APL combines the textual information and application patterns of assignees for classification. We show the performance of Recall and NDCG metrics when the values of K are set to 1, 3, and 5 in \tabref{tab:Ablation_Study}.

From \tabref{tab:Ablation_Study}, we could find that each component contributes to the improvement and our model shows the best performance when it takes all components. In particular, the PSE component captures the unique textual information on patent documents. The ICL component captures the semantic correlation information among IPC codes in taxonomy. The APL component models the publication behavior patterns of each assignee with dual information aggregation.

\subsection{Computational Cost Comparison}
\begin{figure}[!htbp]
    \centering
    \includegraphics[width=1\columnwidth]{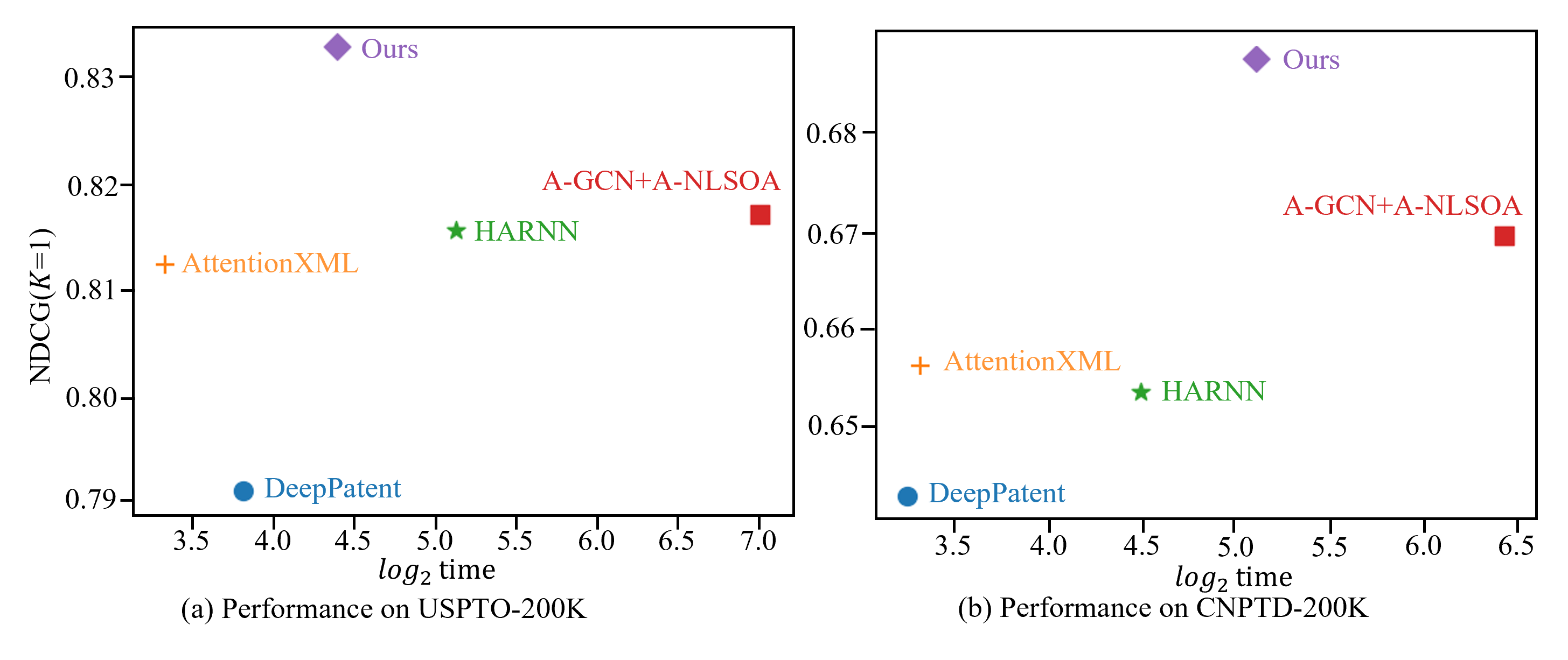}
    \caption{Log-scale evaluation time for various methods on USPTO-200K and CNPTD-200K datasets.}
    \label{fig: runtime_comparison}
\end{figure}
We further provide an analysis of the evaluation time for various methods on USPTO-200K and CNPTD-200K datasets, as shown in \figref{fig: runtime_comparison}. The x-axis of the figure is scaled logarithmically with a base of 2. It is worth noting that the model complexity of our approach primarily stems from the historical pattern modeling component, which is relatively more intricate compared to other baseline methods. However, despite this increased complexity, our model achieves the best performance while maintaining acceptable increments in computational requirements. Therefore, we can conclude that our model strikes a favorable balance between efficiency and effectiveness.

\subsection{Visualization of Hierarchical Embedding}

\begin{figure}[!htbp]
    \centering
    \includegraphics[width=0.65\columnwidth]{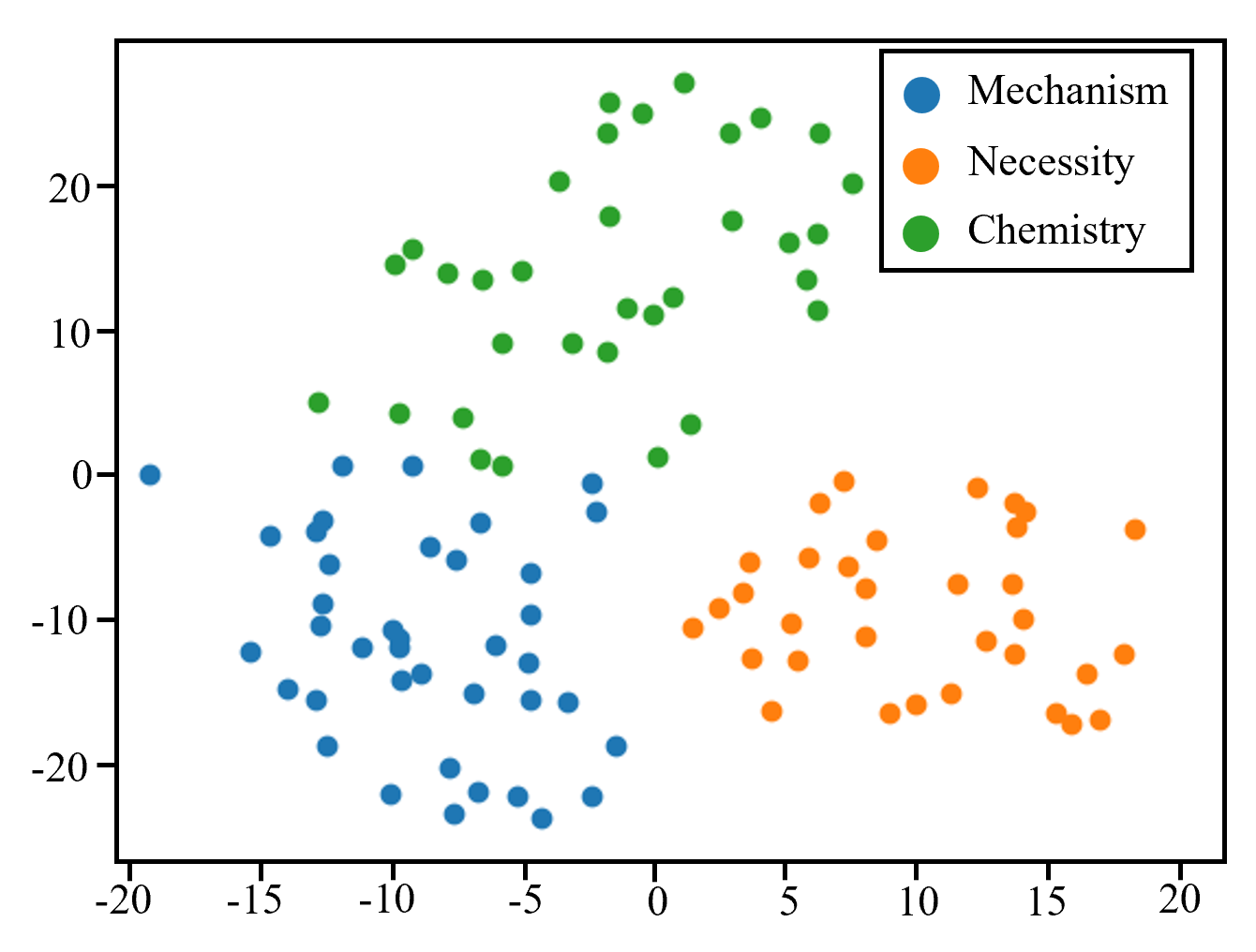}
    \caption{Visualization of embeddings on different fields.}
    \label{fig:Visualization of Embeddings in the third level}
\end{figure}
At last, we show the interpretability of our approach by visualizing the learned embeddings of IPC codes at the third level on USPTO-200K. In this part, we sample three types of IPC codes from a variety of research fields randomly and choose 50 IPC codes in each field randomly. Then we visualize the embeddings of IPC codes with t-SNE \cite{tNSE} in \figref{fig:Visualization of Embeddings in the third level}. We could observe that: 1) the IPC codes in the same field tend to gather more closely than in different fields. Besides, the boundary between the ``Chemistry'' field and the ``Necessity'' field is more distinct than that between the ``Chemistry'' field and the ``Mechanism'' field, which indicates that the semantic relationships are more closely between the ``Mechanism" field and the ``Chemistry'' field. 
It demonstrates that our model captures the semantic correlations among IPC codes. 2) There exist several small clusters in each field while some IPC codes are isolated, which indicates that each IPC code contains unique information in the same field. These findings further illustrate the interpretability of our model, emphasizing its ability to capture meaningful semantic relationships among IPC codes.


%% file: section-6-conclusion.tex
\section{Conclusion}
\label{section-6}
In this paper, we proposed an integrated framework for addressing the patent classification problem, specifically focusing on text descriptions in patent documents, the hierarchical taxonomy of IPC codes, and the historical patent applications of assignees. In contrast to existing works that captured fixed relationships in the hierarchical structure, we provided an IPC codes correlations learning module. This module enabled the adaptive learning of semantic correlations among IPC codes along the taxonomy from both the horizontal and vertical levels by using an attention mechanism. Furthermore, we modeled the higher-order historical publication patterns beyond sequential behaviors for assignees. We accomplished this by utilizing a dual graph neural network that took into account comprehensive information such as textual embeddings and label features and added positional encoding for enhancing sequential information. Through extensive experiments conducted on real-world datasets, we demonstrated the effectiveness of our model and the results highlighted its ability to learn the semantic dependencies among IPC codes.